\documentclass[conference]{IEEEtran}
\IEEEoverridecommandlockouts

\usepackage[table,dvipsnames]{xcolor}
\usepackage{cite}
\usepackage{amsmath,amssymb,amsfonts}
\usepackage{algorithmic}
\usepackage{graphicx}
\usepackage{url}
\usepackage{textcomp}
\usepackage{booktabs}
\usepackage{hyperref}
\usepackage{tabularray}
\usepackage{fontawesome5} 
\usepackage{array}
\usepackage{etoolbox}
\AtBeginEnvironment{thebibliography}{\tiny}

\begin{document}
\title{Revisiting MLLM Token Technology through the Lens of Classical Visual Coding\vspace{-4mm}}

\author{
    Jinming Liu$^{1*}$,
    Junyan Lin$^{1*}$,
    Yuntao Wei$^{1*}$,
    Kele Shao$^{2}$,
    Keda Tao$^{2}$,\\
    Jianguo Huang$^{1}$,
    Xudong Yang$^{1}$,
    Zhibo Chen$^{3}$,
    Huan Wang$^{2}$,
    and Xin Jin$^{1\dagger}$\\
    $^1$Eastern Institute of Technology, Ningbo, China\qquad
    $^2$Westlake University\qquad
    $^3$USTC
\vspace{-5mm}
}


\maketitle
\renewcommand\thefootnote{\fnsymbol{footnote}}
\footnotetext[1]{Equal contribution.}
\footnotetext[2]{Corresponding author: Xin Jin (jinxin@eitech.edu.cn).}

\begin{abstract}
Classical visual coding and Multimodal Large Language Model (MLLM) token technology share the core objective—maximizing information fidelity while minimizing computational cost. Therefore, this paper reexamines MLLM token technology, including tokenization, token compression, and token reasoning, through the established principles of long-developed visual coding area. From this perspective, we (1) establish a unified formulation bridging token technology and visual coding, enabling a systematic, module-by-module comparative analysis; (2) synthesize bidirectional insights, exploring how visual coding principles can enhance MLLM token techniques' efficiency and robustness, and conversely, how token technology paradigms can inform the design of next-generation semantic visual codecs; (3) prospect for promising future research directions and critical unsolved challenges. In summary, this study presents the first comprehensive and structured technology comparison of MLLM token and visual coding, paving the way for more efficient multimodal models and more powerful visual codecs simultaneously.

\end{abstract}


\vspace{-1mm}
\section{Introduction}
\vspace{-1mm}
In the digital media era, image and video coding~\cite{wallace1991jpeg,bross2021vvc,balle2018variational} underpin nearly every visual experience—from real-time video calls and streaming movies to the billions of photos stored in the cloud—by making raw pixel data orders of magnitude smaller while preserving perceptual quality. The core objective is to represent visual content with the minimum number of bits that still meet application-specific fidelity constraints, thereby reducing bandwidth, storage, and latency costs across heterogeneous networks and devices. Classical codecs (here including traditional~\cite{wallace1991jpeg,bross2021vvc} and learning-based neural~\cite{balle2018variational,cheng2020learned,liu2023learned} codecs), typically compress signals by exploiting signal structure through analysis transforms, quantization, and predictive \&
entropy coding, while optimizing explicit rate–distortion (R–D) trade-offs under device and network constraints.

With the rise of deep learning, we are entering an AI-centric era in which applications increasingly depend on multimodal understanding. MLLMs~\cite{bai2025qwen2,li2024llava,chen2024far} extend language models with visual inputs and already support tasks such as captioning, visual question answering, tool use, and embodied reasoning. However, practical deployment is frequently constrained by computational cost: the time and memory complexity of standard attention grow roughly quadratically with the number of tokens~\cite{shang2024llava,bolya2022tome}, so naively encoding rich visual inputs into many tokens may dominate end-to-end latency and energy.

\begin{figure}[t]
    \centering
    \includegraphics[width=1\linewidth]{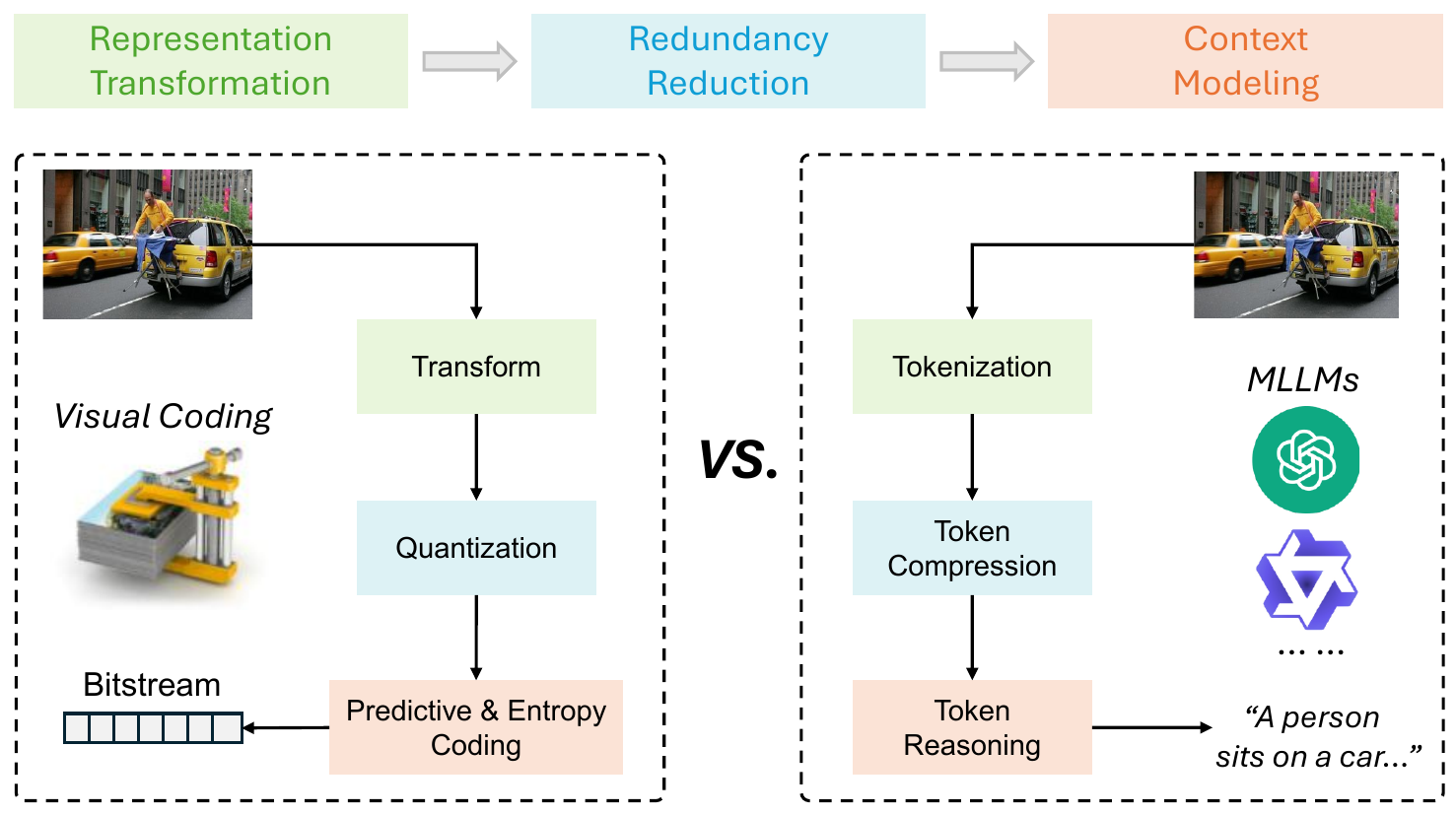}
    \vspace{-7mm}
    \caption{Module-by-Module comparison of visual coding and MLLMs. This paper revisits them from a unified pipeline of ``\textcolor{green!80!black}{\text{Representation Transformation}} $\to$ \textcolor{SkyBlue!90}{\text{Redundancy Reduction}} $\to$ \textcolor{orange!90}{\text{Context Modeling}}''.}
    \vspace{-5mm}
    \label{fig:teaser}
\end{figure}

To address this bottleneck, a growing body of work~\cite{chen2024image,zhang2025vispruner} compresses the visual token stream before or within the MLLM. Techniques include more compact tokenization~\cite{radford2021learning}, token compression~\cite{zhu2024minigpt}, and sparse token reasoning~\cite{chen2024image}. Although developed in different communities, these methods share deep commonality with visual coding. Both pursue compact representations under source constraints and can be cast in a unified R–D view—where ``rate'' is a token/compute budget and ``distortion” measures semantic fidelity (task performance) rather than purely pixel/perceptual error. Thus, this paper revisits MLLM token technology from the perspective of classical visual coding and makes the following contributions:
\begin{itemize}
    \item We present a unified problem formulation that links token technology and visual coding, while comparing each module of them by analogy, as shown in Fig.~\ref{fig:teaser}, to highlight similarities and differences in their underlying philosophies and operational methodologies.
    
    \item We further analyze bidirectional insights from the comparison above and explore how classical coding tools can improve token efficiency and reasoning quality, and how token systems and semantic objectives can guide next-generation semantic codecs.
    \item Finally, we outline future directions and figure out unresolved problems spanning metrics, algorithms, and system design, for both MLLMs and visual codecs.
\end{itemize}

\vspace{-1mm}
\section{Classical Visual Coding and Codecs}
\begin{figure}
    \centering
    \includegraphics[width=1\linewidth]{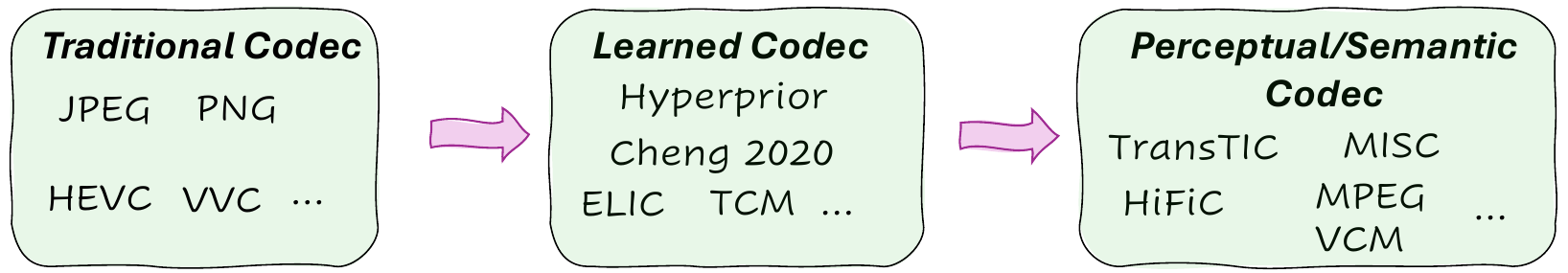}
    \vspace{-6mm}
    \caption{Codec evolution roadmap, from traditional codecs~\cite{wallace1991jpeg,graphics2003specification,bross2013hevc,bross2021vvc} towards more general visual coding~\cite{balle2018variational,he2022elic,cheng2020learned,liu2023learned,chen2023transtic,li2024misc,mentzer2020high,vcm}.}
    \vspace{-4mm}
    \label{fig:coding}
\end{figure}

\vspace{-1mm}
Visual coding aims to represent visual data with minimal bits while retaining crucial visual information, enabling faster transmission and lower storage costs. We analyze the codec evolution roadmap as shown in Fig.~\ref{fig:coding}.

\subsection{Traditional Coding Methods}

Traditional codecs have developed a series of influential standards. These methods typically included the most important three steps: transformation, quantization, and predictive \& entropy coding. For instance, the widely used JPEG~\cite{wallace1991jpeg} standard divides images into 8×8 blocks, applies the Discrete Cosine Transform, and then compresses the coefficients through quantization and entropy coding. PNG~\cite{graphics2003specification} uses filtering and deflate algorithms to maintain exact image fidelity. Advances in video coding, such as HEVC~\cite{bross2013hevc} and VVC~\cite{bross2021vvc}, have brought more sophisticated block partitioning, prediction, and transform techniques, improving performance. 

\subsection{Learning-based Coding Methods}
The resurgence of artificial neural networks has driven rapid progress in learned image compression~\cite{he2022elic}, now surpassing even the traditional standards like VVC~\cite{bross2021vvc} in RD performance by utilizing end-to-end optimization. For example, Google~\cite{balle2018variational,minnen2018joint} proposed hyperprior models, which use side information to estimate better latent code distributions. Some works~\cite{cheng2020learned,liu2023learned} introduced better transformations, achieving results comparable to VVC~\cite{bross2021vvc}. DCVC-series~\cite{jia2025towards} also showed superior performance on video compared to traditional standards, while achieving over 100 FPS for 1080p video on a GPU.

\subsection{Perceptual/Semantic-driven Coding Methods}

Recently, image compression is moving beyond pixel-level fidelity to address either perceptual quality for humans or semantic utility for machines~\cite{chen2023transtic}. Perceptual codecs aim to optimize images for human perception. These methods relied on GAN-based models~\cite{agustsson2019generative, mentzer2020high} or diffusion models~\cite{mao2024perceptual,xu2025picd} to produce visually pleasing results. In parallel, semantic codecs~\cite{li2024misc,liu2022semantic,chen2023transtic} focus on preserving information needed for machine tasks such as classification and detection, allowing better rate-accuracy. Jointly optimizing the compression model and task network~\cite{li2024human} has been shown to improve machine task accuracy. Some works~\cite{liu2024rate, jin2023semantical} enabled scalable bitstreams that can flexibly serve both human and machine needs. Reflecting these trends, industry and standardization bodies have launched initiatives like MPEG Video Coding for Machines (VCM)~\cite{vcm} and JPEG AI~\cite{jpegai} to unify perceptual and semantic goals within a single framework.

\section{MLLM Token Technology}
An MLLM’s token working flow has three main stages: the tokenizer first extracts tokens, a post-processing step (e.g., token compression), and the resulting tokens are fed into LLMs for reasoning, completing different downstream tasks.

\subsection{Visual Tokenizers in MLLMs}
Visual tokenizers in MLLMs convert image inputs into structured token sequences. 
They directly influence token efficiency and cross-modal alignment, can be categorized into two classes as follows:

\subsubsection{Contrastive learning-based visual tokenizers}

Contrastive tokenizers dominate mainstream MLLMs by producing semantically rich image representations. CLIP~\cite{radford2021learning} is a foundational model that aligns image and text embeddings through large-scale contrastive pretraining. Building on CLIP, SigLIP~\cite{zhai2023sigmoid} replaces the contrastive loss with a sigmoid-based variant and removes the [CLS] token, enabling denser supervision via averaged patch features. 

\subsubsection{Self-Supervised Visual Tokenizers}
In contrast, self-supervised tokenizers like DINO~\cite{caron2021emerging} require no text supervision. Using a teacher–student framework, DINO promotes view consistency, leading to emergent semantic structure in intermediate tokens. DINOv2~\cite{oquab2024dinov2} scales this approach with improved alignment and stronger performance on linear classification and segmentation.

\begin{figure}[ht]
    \centering
    \vspace{-3mm}
    \includegraphics[width=0.8\linewidth]{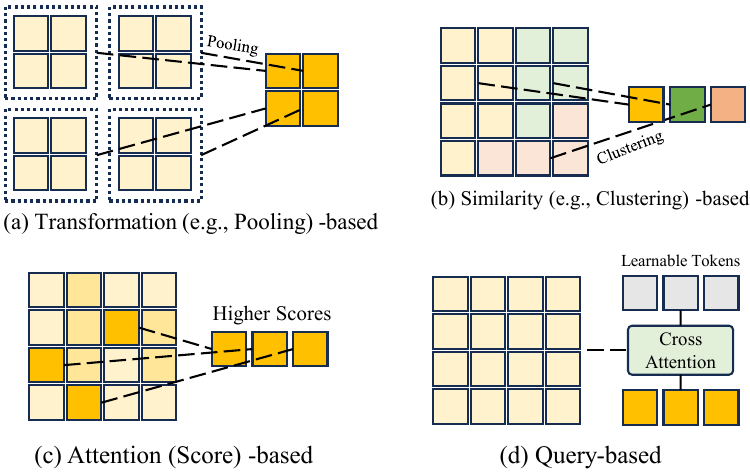}
    \vspace{-3mm}
    \caption{Comparison of different token compression strategies.}
    \vspace{-3mm}
    \label{fig:comparision}
\end{figure}

\subsection{Token Compression Strategies in MLLMs}

High-resolution images often contain many redundant or uninformative tokens. To address this, researchers have proposed various token compression strategies for image modalities\cite{shao2025tokens}, which can be broadly categorized into four types: transformation-based, similarity-based, attention-based, and query-based methods as shown in Fig. \ref{fig:comparision}.

\subsubsection{Transformation-based Compression}

These methods~\cite{wang2024qwen2,chen2024far,li2024llava,chu2023mobilevlm} compress image tokens by applying spatial transformations such as pixel unshuffle, pooling, or convolution, effectively reducing token count while preserving local structure. These lightweight operations are typically applied at the model frontend, requiring minimal computation.

Representatively, Qwen2~\cite{wang2024qwen2} and InternVL~\cite{chen2024far} series adopt pixel unshuffle to convert high-resolution images into lower-resolution but higher-channel representations, followed by an MLP for feature dimension alignment. LLaVA-OneVision~\cite{li2024llava} uses average pooling or bilinear interpolation to learn multi-granular visual representations for adaptive reasoning.
Unlike those approaches that implement non-parametric operations, MobileVLM~\cite{chu2023mobilevlm} employs depthwise convolution to achieve over 75\% token reduction while retaining spatial locality.

\subsubsection{Similarity-based Compression}
These methods~\cite{bolya2022tome,chai2024auroracap} reduce the token count by merging redundant or semantically similar visual tokens into representative ones, making them particularly effective for highly redundant image inputs. A foundational work is ToMe~\cite{bolya2022tome}, which accelerates ViT inference by inserting token clustering modules into each Transformer layer and fusing tokens based on bidirectional cosine similarity.
AuroraCap~\cite{chai2024auroracap} adopts this strategy and demonstrates its effectiveness in MLLMs by progressively merging tokens after each visual layer.

\subsubsection{Attention-based Compression}
Attention-based strategies~\cite{chen2024image},~\cite{shang2024llava,zhang2025vispruner,xing2024pyramiddrop,shao2025holitom} leverage the inherent sparsity in attention distributions to dynamically prune less important tokens, thereby improving inference efficiency. These methods can be grouped into two categories based on the pruning stage.

\textbf{Tokenizer Stage:} pruning is guided by attention scores computed within the vision encoder, typically with respect to the [CLS] token. These scores are used to evaluate token importance. For example, PruMerge~\cite{shang2024llava} selects high-attention tokens as cluster centers and aggregates the rest via KNN-based merging. VisPruner~\cite{zhang2025vispruner} retains salient tokens based on attention and further ensures token diversity through iterative semantic deduplication, preserving both relevance and variety.

\textbf{LLM Stage:}
pruning is guided by attention scores within the LLM. This enables more semantically aware token selection, crucial for downstream understanding tasks. FastV~\cite{chen2024image} observes that visual tokens receive minimal attention beyond the early layers and proposes to prune half of the visual tokens immediately following the second layer, achieving a significant speed-up with negligible performance loss. PyramidDrop~\cite{xing2024pyramiddrop} extends this idea by applying layer-wise adaptive pruning based on attention distributions, striking a better balance between efficiency and accuracy.

\subsubsection{Query-based Compression}
In many multimodal tasks, such as visual question answering, a large portion of visual information may be irrelevant to the task objective. Query-based compression methods~\cite{li2023blip,zhu2024minigpt} aim to reduce this redundancy by using learnable query tokens or external prompts to extract compact visual representations that are most useful for downstream language processing.
The BLIP series offers a representative solution. BLIP-2~\cite{li2023blip} uses fixed learnable query tokens to distill visual features into a fixed-length representation aligned with the language model, though it does not condition on task instructions. 

MiniGPT-4~\cite{zhu2024minigpt} follows similar designs, where visual features are distilled via learnable queries and projected into compact embeddings, enabling modular and efficient multimodal reasoning.

\definecolor{tableheadcolor}{gray}{0.92}

\begin{table*}[h]
\centering
\caption{A Comparative Overview of Visual Coding and MLLM Token Technology.}
\vspace{-2mm}
\label{tab:comparison}
\renewcommand{\arraystretch}{1.6} 
\scriptsize
\begin{tabular}{>{\bfseries}p{2.0cm} p{6.8cm} p{7cm}}
\toprule
\rowcolor{tableheadcolor}
\textbf{Aspect} & \textbf{Classical Visual Coding} & \textbf{MLLM Token Technology} \\
\midrule
Primary Goal & Efficient data storage and transmission. & Efficient machine computation and reasoning. \\
\hline
Core Optimization (Rate-Distortion) & 
\textbf{Rate:} Transmission cost (bits). \newline \textbf{Distortion:} Fidelity error for \textit{human observers} (e.g., PSNR). &
\textbf{Rate:} Computational cost (token count). \newline \textbf{Distortion:} Semantic error for \textit{machine observers} (task performance). \\
\hline
Core Pipeline &
\textbf{1. Transformation:} Transforms for \textit{energy compaction}. \newline
\textbf{2. Redundancy Reduction:} Quantization guided by \textit{information distribution}. \newline
\textbf{3. Context Modeling:} Predictive and entropy coding to \textit{reduce bits}. &
\textbf{1. Transformation:} Vision encoders for \textit{semantic extraction}. \newline
\textbf{2. Redundancy Reduction:} Token pruning/merging/VQ guided by \textit{semantic relevance}. \newline
\textbf{3. Context Modeling:} Auto-regressive process to \textit{construct meaning}. \\
\hline
Bidirectional Insights &
\textbf{Influences MLLMs by:} Providing principles for token reduction (RDO, VQ); Variable-length tokenization; preprocessing method. &
\textbf{Influences Codecs by:} Semantic prior for rate control; enabling generative compression; and offering a path to universal compressors. \\
\bottomrule
\end{tabular}
\vspace{-5mm}
\end{table*}
\section{Bridging Visual Coding and MLLM Tokens}

\subsection{The Unified Objective: A Rate-Distortion Perspective}
The optimization challenge in both fields is to solve the rate-distortion trade-off, formally expressed as minimizing the cost function $J = R + \lambda D$~\cite{cheng2020learned,balle2018variational}. While the formula is shared, its components are defined by the end-user. For \textbf{classical visual coding}~\cite{cheng2020learned,balle2018variational}, the goal is to serve the human eye: \textbf{Rate ($R$)} is the \textit{transmission cost} measured in bits, and \textbf{Distortion ($D$)} is the \textit{fidelity error} (e.g., PSNR, MS-SSIM). Conversely, for \textbf{MLLM token technology}~\cite{li2024llava}, the goal is to serve the machine model: \textbf{Rate ($R$)} becomes the \textit{computational cost} dictated by the number of tokens, and \textbf{Distortion ($D$)} is the \textit{semantic error} measured in task performance degradation. The unified problem, therefore, is to find an optimal representation by balancing cost and information loss, tailored either for human perception or machine cognition.

\subsection{A Unified Pipeline: A Modular-based Analogy}
To solve their respective rate-distortion problems, both systems independently converged on functionally analogous pipelines as shown in Fig.~\ref{fig:teaser} and Table~\ref{tab:comparison}. We can analyze this pipeline by its core stages.

\subsubsection{Stage 1: Representation Transformation to a Compact Representation}
The initial pipeline stage in both representation transforms the pixels into a compact representation, but the nature of this compactness is fundamentally different: Visual coding pursues \textbf{information compactness}~\cite{liu2023learned,balle2018variational}. It employs handcrafted transforms, like the DCT, or learned transforms to de-correlate pixels as compact representations, optimized for bit-saving compression algorithms while aiming to preserve fidelity. Conversely, MLLM tokenization aims for \textbf{semantic compactness}~\cite{radford2021learning,zhai2023sigmoid}. It uses learned semantic encoders to extract visual meaning into tokens, resulting in a representation that is semantically dense and optimized for the cognitive efficiency of a downstream reasoning model.

\subsubsection{Stage 2: Redundancy Reduction}
Following stage 1, both pipelines apply a post-processing stage to control representation cost by discarding less important content. In visual coding, quantization is often guided by hand-crafted rules~\cite{wallace1991jpeg} or distribution statistics~\cite{gadot2025rl}—either dropping low-salience details outright or assigning them fewer bits. Likewise, MLLMs prune tokens produced by the vision encoder using hand-set rules or importance/similarity measures~\cite{bolya2022tome,shang2024llava}, allocating more tokens to important information and fewer to the rest. Although the mechanisms differ, both are fundamentally rate-control strategies that filter the representation according to a guiding principle: the former optimizes for human-perceived fidelity, the latter for machine-perceived relevance.

\subsubsection{Stage 3: Context Modeling}
Finally, both systems leverage context to enhance efficiency, but their predictive mechanisms serve different ends. Visual coding employs explicit predictive coding in inter/intra-prediction~\cite{bross2021vvc}. It uses previously encoded data (spatial or temporal neighbors) to predict the current information unit and then encodes only the prediction residual.  Also, some entropy models~\cite{he2022elic} in visual coding use context models to help distribution estimation for better bits estimation. They are strategies purely aimed at maximizing compression efficiency by exploiting statistical redundancy. In parallel, the MLLM architecture inherently performs prediction through its auto-regressive nature~\cite{bai2025qwen2}. The model leverages the entire preceding sequence of tokens (both text and visual) as context to predict the next token. This predictive process is not for compressing a residual but for ensuring semantic coherence in its generated output. Thus, while both use context to inform the next step, one predicts to reduce bits, the other predicts to construct meaning.

\subsection{Bidirectional Synergy and Future Co-evolution}
\subsubsection{How Visual Coding Principles Can Refine MLLMs}
\paragraph{Transform-based Pre-processing} Inspired by classical codecs, MLLM tokenization can be improved by first transforming images into a more compressible domain. This includes using some transforms to operate in the frequency domain~\cite{feng2024docpedia,pertsch2025fast}, or adopting principles from video coding to separately tokenize ``keyframe" information and more redundant ``motion-like" information, leading to a more efficient initial representation~\cite{gadot2025rl}.

\paragraph{Principled Redundancy Reduction} Some compression algorithms provide direct inspiration for token reduction. The concept of Run-Length Coding, for instance, can be adapted to merge sequences of similar tokens~\cite{choudhury2024don}. More broadly, ``information-preserving guided selection" strategies, which prune or merge tokens based on their contribution to overall information, are a direct application of rate-control philosophies to the semantic domain~\cite{tan2025tokencarve}.

\paragraph{Theoretical Foundations} Drawing from the ultimate theoretical principle of compression—Kolmogorov Complexity and its computable proxy, the Minimum Description Length (MDL)—inspires a shift towards variable-length tokenization~\cite{duggal2025single}. This allows models to represent simple images with fewer tokens and complex ones with more, achieving a principled trade-off between token count and semantic fidelity, akin to a quality parameter in visual coding.

\paragraph{Vector Quantization} Vector Quantization (VQ), a cornerstone of signal processing, has been successfully imported into MLLMs~\cite{geng2025x}. By mapping continuous embeddings to a discrete, learned codebook, VQ not only reduces the representation cost but also significantly improves the quality and stability of generative tasks.

\subsubsection{How MLLM Intelligence Can Refresh Codecs}
\paragraph{Semantic Prior}
 An MLLM can act as a high-level ``brain" for a codec~\cite{liu2024tell,li2024misc}. By first providing a semantic analysis of the scene, it can generate a priority map that guides the codec to dynamically allocate more bits to critical regions (e.g., faces, text) while aggressively compressing unimportant backgrounds. This optimizes for downstream task performance rather than uniform perceptual quality.

\paragraph{Token Coding} A paradigm shift is emerging where the target of compression is no longer pixels, but the semantic tokens themselves. Compressing tokens~\cite{gao2024feature,qiao2025token} directly yields superior performance for machine-vision tasks. 
\paragraph{Generative Compression}An MLLM can distill an image into a compact text description, which is then transmitted and used by a generative model to reconstruct the image, enabling ultra-low bitrates.

\paragraph{Universal Predictive Compressor} At a fundamental level, the powerful next-token prediction capability of large models makes them extraordinarily effective general-purpose compressors. Large models have already demonstrated the ability to compress general data streams~\cite{li2025lossless,deletanglanguage}, surpassing specialized codecs like PNG. This positions MLLMs not merely as a tool for compression, but as the potential core engine for future universal data compression standards.

\subsubsection{System Perspective}
The convergence of MLLM technology and visual coding naturally leads to a practical system architecture: an edge–cloud co-design. The core challenge in deploying large-scale MLLMs for real-world tasks is the bottleneck between the data source (edge) and the processing center (cloud), driven by two constraints: limited network bandwidth and the high computational cost of cloud-based MLLMs. Several studies~\cite{kaobridging,li2024high} have explored MLLM-oriented visual coding that reduces transmission bandwidth while preserving MLLM performance. Some works place the tokenizer at the edge~\cite{qiao2025token}, encode and transmit tokens directly without image reconstruction at the decoder. Nonetheless, jointly optimizing the bitrate–complexity–performance trade-off remains an open problem.

\section{Future Directions and Open Problems}
The convergence of visual coding and MLLM token technology opens a rich frontier for research. On the codec side, MLLM tokens can serve as semantic priors to enable object-/region-level adaptive bit allocation, language-/task-driven temporal rate control, scalable semantic layering (transmitting essentials first, details later), generative completion with ultra-low bitrates, and edge-side lightweight semantic summaries. Conversely, importing coding principles into token systems suggests variable-length and hierarchical semantic tokenization (base summaries with enhancement details), better video-stream representations, spatiotemporal prediction with redundancy suppression, and information-theoretic foundations for token semantic entropy. At the system level, edge–cloud co-design for semantic rate control and efficient MLLM inference, together with hardware–software co-design of accelerators under energy/latency constraints, constitute pressing directions for more general, scalable, reliable semantic communication.

\begingroup
  \let\origfootnotesize\footnotesize   
  \renewcommand{\footnotesize}{\scriptsize}
  \bibliographystyle{IEEEtran}
  \bibliography{references}
\endgroup
\end{document}